\documentclass{bmvc2k}

\usepackage{amsmath}
\usepackage{amssymb}
\usepackage{multirow}

\title{Accurate and Compact Convolutional Neural Networks with Trained
	Binarization}

\addauthor{Zhe Xu}{zhexu22-c@my.cityu.edu.hk}{1}
\addauthor{Ray C.C. Cheung}{r.cheung@cityu.edu.hk}{1}

\addinstitution{
	Department of Electrical Engineering\\
	City University of Hong Kong\\
	Hong Kong SAR, China
}

\runninghead{Z. Xu, R. Cheung}{Accurate networks with Trained Binarization}

\def\etal{\emph{et al}\bmvaOneDot}
\def\ie{\emph{i.e}\bmvaOneDot}

\begin{document}
	
	\maketitle
	
	\begin{abstract}
		Although convolutional neural networks (CNNs) are now widely 
		used in various computer vision applications, its huge resource
		demanding on parameter storage and computation makes the
		deployment on mobile and embedded devices difficult.  Recently,
		binary convolutional neural networks are explored to help
		alleviate this issue by quantizing both weights and activations
		with only 1 single bit. However, there may exist a noticeable
		accuracy degradation when compared with full-precision models.
		In this paper, we propose an improved training approach towards
		compact binary CNNs with higher accuracy. Trainable scaling
		factors for both weights and activations are introduced to
		increase the value range. These scaling factors will be trained
		jointly with other parameters via backpropagation. Besides, a
		specific training algorithm is developed including tight
		approximation for derivative of discontinuous binarization
		function and $L_2$ regularization acting on weight
		scaling factors. With these improvements, the binary CNN
		achieves 92.3\% accuracy on CIFAR-10 with VGG-Small network. On
		ImageNet, our method also obtains 46.1\% top-1 accuracy with
		AlexNet and 54.2\% with Resnet-18 surpassing previous works.   
		
	\end{abstract}

	\section{Introduction}
	\label{sec:intro}
	Convolutional neural networks (CNNs) have achieved great success in a
	wide range of real-world applications such as image
	classification~\cite{Krizhevsky_2012, Simonyan_2015, He_2016}, object
	detection~\cite{Girshick_2014, Redmon_2016, Liu_2016}, visual relation
	detection~\cite{Zhang_2017, Zhang_2017ppr} and image style
	transfer~\cite{Gatys_2016, Chen_2017} in recent years. However, the
	powerful CNNs are also accompanied by the large model size and high
	computational complexity. For example, VGG-16~\cite{Simonyan_2015}
	requires over 500MB memory for parameter storage and 31GFLOP for a
	single image inference.  This high resource demanding makes it
	difficult to deploy on mobile and embedded devices.
	
	To alleviate this issue, several kinds of approaches have been
	developed to compress network and reduce computational cost. The first
	approach is to design compact network architectures with similar
	performance. For example, SqueezeNet~\cite{Iandola_2016} utilized 1x1
	convolution layer to squeeze the output channel number before
	computationally expensive 3x3 convolution operations. More recently,
	MobileNet~\cite{Howard_2017} and ShuffleNet~\cite{Zhang_2018}
	introduced efficient depth-wise and group convolution to replace the
	normal complex convolution operation. The second approach is to reduce
	or compress network parameters. \cite{Lebedev_2015}
	and~\cite{Novikov_2015} achieved to compress network weights using
	tensor decomposition. Connection pruning was employed
	in~\cite{Han_2015} to reduce parameters by up to 13 times for AlexNet
	and VGG-16. The third category is to quantize network parameters
	presented in~\cite{Zhuang_2018, Dong_2017, Lin_2016, Zhu_2017,
		Zhou_2017, Hou_2018}. Network quantization can reduce memory
	requirement efficiently because parameter values are represented with
	less bits.  At the same time, it can alleviate the computational cost
	issue since floating-point calculations are transferred into fixed-point
	calculations with less computation resources.
	
	Furthermore, as an extreme case of parameter quantization, binary
	convolutional neural networks quantize both weights and activations
	with 1 bit~\cite{Hubara_2016, Rastegari_2016}. It has attracted large
	research interests because binary CNNs can reduce network size by 32
	times compared with full precision and replace the multiplication 
	with bitwise logic operations. As a result, binary CNNs are suitable 
	for accelerator implementations on hardware platforms such as 
	FPGA~\cite{Umuroglu_2017}.  However, network binarization may decrease 
	accuracy noticeably due to extremely limited precision. It is still 
	a challenge and needs further exploration towards better inference 
	performance. 
	
	In this paper, an improved training approach for binary CNNs is proposed
	which is easy to be implemented on hardware platforms. Our approach
	includes three novel ideas:
	
	\begin{enumerate}
		\item Trainable scaling factors are introduced for weight and activation
		binarization. Previous works such as XNOR-Net~\cite{Rastegari_2016} set the mean
		value of weights as the scaling factor, however it results in minimum
		quantization error but cannot ensure the best inference accuracy. Instead, we
		employ the trainable scaling factors for both weights and activations and update
		them jointly with other network parameters. 
		\item Derivative approximation is discussed for binary network training. Since 
		the derivative of binarization function is like an impulse function, it
		is not suitable for backpropagation. We propose to use a higher order
		approximation for weight binarization and a long-tailed approximation for
		activation binarization as a trade-off between tight approximation and smooth
		backpropagation.
		\item The $L_2$ regularization term is now acting on the weight scaling
		factors directly. In our approach, weight scaling factors represent the actual 
		binary filters, the $L_2$ regularization should be modified accordingly for 
		better generalization capability.  
	\end{enumerate}
	
	The proposed binary network approach achieves better inference accuracy
	and faster convergence speed. The rest of this paper is organized as
	follows.  Section~\ref{sec:related_work} reviews previous related
	works. The proposed binary network training approach is introduced in
	detail in Section~\ref{sec:method}. Experimental results are provided
	in Section~\ref{sec:results}. Finally, Section~\ref{sec:conclusion}
	gives the conclusion. 
	
	\section{Related Work}
	\label{sec:related_work}
	Previous works~\cite{Zhuang_2018, Dong_2017, Lin_2016, Zhu_2017,
		Zhou_2017, Hou_2018} already demonstrated that quantization can reduce much
	memory resources and computational complexity for various CNN structures. One
	extreme case of quantization is to constrain real value with only 1
	bit,~\ie binarization. It can be further divided into two
	subcategories: one is only binarizing weights and leaving activations
	full-precision or quantized, another is binarizing both weight and
	activation values.
	
	\textbf{Weight-only binarization methods:} 
	Courbariaux~\etal~\cite{Courbariaux_2015} firstly proposed to train networks
	constraining weights to only two possible values, -1 and +1. They introduced a
	training method, called BinaryConnect, to deal with binary weights in both
	forward and backward propagations and obtained promising results on small
	datasets. 
	In~\cite{Cai_2017}, Cai~\etal binarized weights and proposed a half-wave
	Gaussian quantizer for activations. The proposed quantizer exploited the
	statistics of activations and was efficient for low-bit quantization.
	Later Wang~\etal~\cite{Wang_2018} further extended~\cite{Cai_2017}'s idea with sparse
	quantization. Besides, they proposed a two-step quantization framework: code
	learning and transformation function learning. In code learning step, weights
	were of full-precision and activations were quantized based on Gaussian distribution
	and sparse constraint. Then the weights binarization was solved as a non-linear
	least regression problem in the second step.  
	In~\cite{Hu_2018}, Hu~\etal proposed to transfer binary weight networks training
	problem into a hashing problem. This hashing problem was solved with
	alternating optimization algorithm and the binary weight network was then
	fine-tuned to improve accuracy.
	
	\textbf{Complete binarization methods:} Although weight-only binarization
	methods already save much memory resources and reduce multiplications,
	further binarization on activation can transfer arithmetic to bit-wise logic
	operations enabling fast inference on embedded devices. 
	As far as our knowledge goes,~\cite{Hubara_2016} was the first work binarizing
	both weights and activations to -1 and +1. The work obtained 32 times
	compression ratio on weights and 7 times faster inference speed with comparative
	results to BinaryConnect on small datasets like CIFAR-10 and SVHN. However,
	later results showed that this training method was not suitable for large
	datasets with obvious accuracy degradation. 
	Later Rastegari~\etal~\cite{Rastegari_2016} proposed XNOR-Net to improve the
	inference performance on large-scale datasets. It achieved better trade-off
	between compression ratio and accuracy in which scaling factors for both weights
	and activations were used to minimize the quantization errors.   
	DoReFa-Net~\cite{Zhou_2016} proposed by Zhou~\etal inherited the idea of
	XNOR-Net and provided a complete solution for low-bit quantization and binarization of
	weights, activations and gradients.
	
	In~\cite{Tang_2017}, Tang~\etal explored several strategies to train
	binary neural networks. The strategies included setting small learning
	rate for binary neural network training, introducing scaling factors for
	weights using PReLU activation function and utilizing regularizer to
	constraint weights close to +1 or -1. They showed the binary neural
	networks achieved similar accuracy to XNOR-Net with simpler training
	procedure.  In~\cite{Lin_2017}, Lin~\etal
	proposed ABC-Net towards accurate binary convolutional neural
	network. Unlike other approaches, they proposed to use the
	linear combination of multiple binary weights and binary activations
	for better approximation of full-precision weights and activations.
	With adequate bases, the structure could finally achieve close results
	to full-precision networks. In recent work~\cite{Liu_2018}, Liu~\etal
	proposed Bi-Real Net in which the real activations are added to the
	binary activations through a shortcut connection to increase the
	representational capability. Moreover, they provided a tight
	approximation of sign function and proposed to pre-train full-precision
	neural networks as the initialization for binary network training.   
	
	\section{Proposed Binary Network Training}
	\label{sec:method}
	
	Training an accurate binary CNN has two major challenges:
	one is the extremely limited value range because of binary data,
	another is the difficult backpropagation in training procedure caused
	by the derivative of binarization function. In this section, the
	proposed binary CNN training approach is introduced in order to address
	the above two issues.
	
	\subsection{Binarization with Trainable Scaling Factors}
	
\begin{figure*}[!t]
		\centering
		\includegraphics[width = 5in]{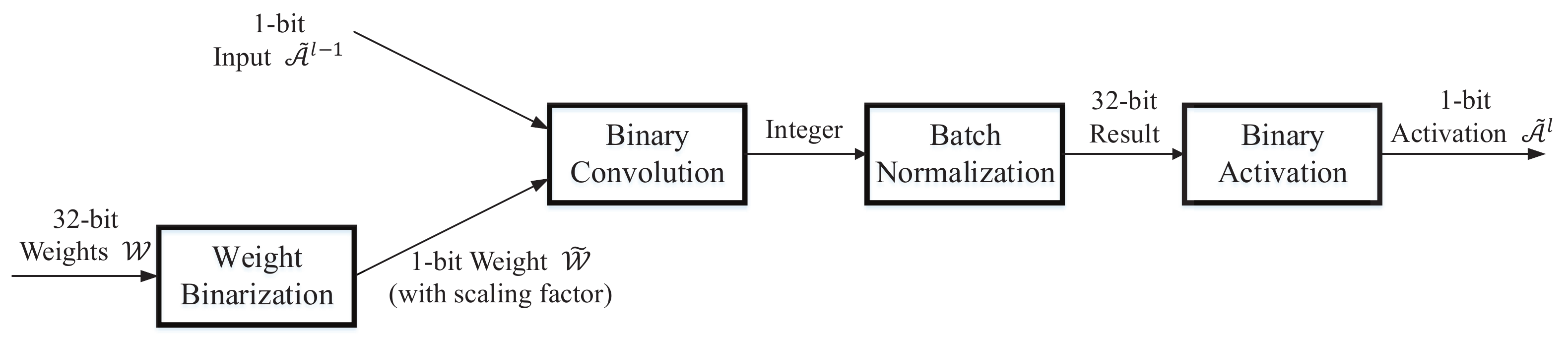}
		\caption{Forward process in binary convolution layer.}
		\label{fig:Binarization}
	\end{figure*}

	We first briefly review the binarization operation in one convolution layer,
	which is shown in Figure~\ref{fig:Binarization}. As binary weighs are difficult
	to update with gradient-based optimization methods due to the huge gap between
	different values, it is common to reserve full-precision weights during training
	process. The binary weights are then obtained from real values via the
	binarization function. The input of the convolution operation is actually the 
	activation output $\tilde{\mathcal{A}}$ of the previous layer, the binary 
	convolution is represented as
	
	\begin{equation}
	z = \tilde{\mathcal{A}} * \tilde{\mathcal{W}}
	\end{equation}
	Since the input $\tilde{\mathcal{A}}$ and weights $\tilde{\mathcal{W}}$ are both
	binary, the convolution operation can be implemented with bitwise logic and
	$popcnt$ operations to get integer results $z$ similar as~\cite{Rastegari_2016}.
	After batch normalization, the integer results $z$ become real values within a
	certain range and they are binarized in the activation step to generate binary
	output feature map for the next convolution layer.
	
	We use a sign function $sgn(x)$ and a unit step function $H(x)$ for weight
	and activation binarization, respectively. $sgn(x)$ and $H(x)$ are 
	
	\begin{equation}
	sgn(x)=\begin{cases}
	1,  & x \ge 0 \\
	-1, & x < 0
	\end{cases}
	\qquad	\qquad
	H(x)=\begin{cases}
	1,  & x \ge 0 \\
	0, & x < 0
	\end{cases}
	\end{equation}
	
	However, $sgn(x)$ and $H(x)$ highly restrict the results to be only
	three possible values $\{-1,0,+1\}$. To increase the value range,
	\cite{Rastegari_2016} and~\cite{Zhou_2016} proposed to use the average
	of absolute values of each output channel as scaling factors for
	weights. This minimizes the binarization errors but does not ensure the
	optimal inference performance. Instead we propose to set trainable
	scaling factors directly and these scaling factors are updated through
	backpropagation as part of the network parameters.
	
	Given the real value weight filter $\mathcal{W}\in
	\mathbb{R}^{c_{o}\times c_i\times k\times k}$ where $c_o$ and $c_i$
	stand for output and input channels respectively, we set a scaling
	factor $\alpha \in \mathbb{R}^{c_{o}}$. Then the weight binarization is
	represented as 
	
	\begin{equation}
	\tilde{\mathcal{W}}_i = \alpha_i \cdot sgn(\mathcal{W}_i)
	\end{equation}
	$\tilde{\mathcal{W}}_i$ stands for the binary weight at output channel $i$.
	Importing the scaling factor $\alpha$ enables binary weights
	$\tilde{\mathcal{W}}$ to have different value magnitudes in each output channel.
	
	Similarly, we set the scaling factor $\beta$ for binary activation
	$\tilde{\mathcal{A}}\in\mathbb{R}^{c_{o}\times w\times h}$
	\begin{equation}
	\tilde{\mathcal{A}}_i = \beta \cdot H(\mathcal{A}_i-\tau_i)
	\end{equation}	
	We set a threshold $\tau \in \mathbb{R}^{c_o}$ as a shift parameter
	trying to extract more information from the original full-precision
	$\mathcal{A}$. By importing parameter $\tau$, only important values will be activated and other small values are set to be zero. It should be noted that the shift parameter $\tau$ will also be updated during network training. $\beta$ is identical for a single activation layer to make it compatible with hardware implementation which will be discussed in Section~\ref{subsec:hardware}.
	
	\subsection{Training Algorithm}
	
	\begin{figure*}[!t]
		\centering
		\includegraphics[width = 4in]{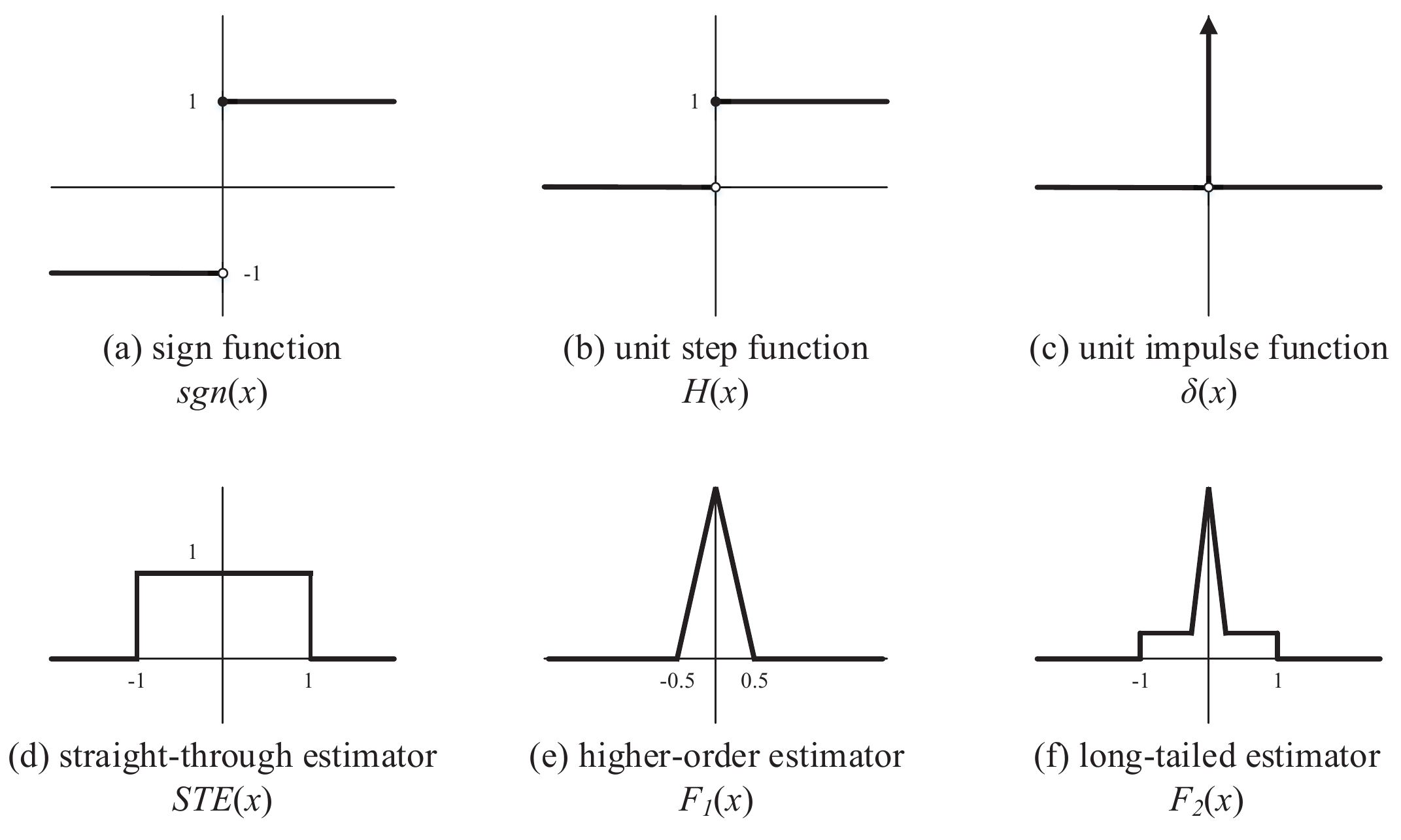}
		\caption{Different approximations for derivatives of $sgn(x)$ and $H(x)$.}
		\label{fig:approxiamtion}
	\end{figure*}
	
	As we discussed before, $sgn(x)$ and $H(x)$ are used to binarize
	weights and activations, respectively. They are illustrated in
	Figure~\ref{fig:approxiamtion}(a) and~\ref{fig:approxiamtion}(b). To
	update network parameters, their derivatives are required. However as
	we can see in Figure~\ref{fig:approxiamtion}(c), their derivatives are
	like an impulse function whose value is zero almost everywhere and
	infinite at zero. Thus it cannot be applied for backpropagation and
	parameter updating directly.  In~\cite{Hubara_2016, Rastegari_2016,
		Zhou_2016}, a clip function was employed to approximate the binarization
	functions. The derivate of $sgn(x)$ is represented in Eq.~\ref{eq:STE} based 
	on straight through estimator (STE)~\cite{Bengio_2013} and it is illustrated 
	in Figure~\ref{fig:approxiamtion}(d).
	
	\begin{equation}
	\label{eq:STE}
	\frac{d}{dw}sgn(x) \approx \mathbf{1}_{\{|w| \le 1\}}
	\end{equation}
	
	Instead of Eq.~\ref{eq:STE}, \cite{Liu_2018} utilized a piecewise polynomial
	function as a tighter approximation. In this paper, we further develop this 
	higher-order estimator approach. For $sgn(x)$ used in weight binarization, we give
	derivative approximation $F_1(x)$ whose active range is between $[-0.5,0.5]$
	illustrated in Figure~\ref{fig:approxiamtion}(e). It is a piecewise linear
	function.
	
	\begin{equation}
	\label{eq:F1}
	\frac{d}{dw}sgn(x) \approx F_1(x) =
	\begin{cases}
	4-8|x|,  & -0.5 \le x \le 0.5 \\
	0,       & otherwise 
	\end{cases}
	\end{equation}
	
	For the derivative of $H(x)$, an overtight approximation may
	not be an optimal choice because it will affect the backpropagation to
	shallow layers far away from the network output. For an overtight
	approximation, the gradient value will be near zero in a large scope
	resulting in gradient vanishing. To alleviate this problem, we use a
	long-tailed higher-order estimator $F_2(x)$ whose active range is
	between $[-1,1]$, shown in Figure~\ref{fig:approxiamtion}(f). It is a
	piecewise function with tight approximation near zero and small
	constant value in a wider range.
	
	\begin{equation}
	\label{eq:F2}
	\frac{d}{dw}H(x) \approx F_2(x) =
	\begin{cases}
	2-4|x|,  & -0.4 \le x \le 0.4 \\
	0.4,     & 0.4 < |x| \le 1 \\
	0,       & otherwise 
	\end{cases}
	\end{equation}
	
	Based on Eq.~\ref{eq:F1} and~\ref{eq:F2}, the backpropagation can be
	performed smoothly during binary network training. It should be noted
	that there certainly exist other estimators as a trade-off between
	tight approximation and smooth backpropagation, such as Swish-like
	function~\cite{Elfwing_2018}. In this section, we provide a simple yet
	efficient approximation and leave other options as our future work. 
	
	\subsection{Regularization on Scaling Factors}
	
	Since deep CNNs usually have a tremendous parameter set, a good regularization term
	is necessary during training process for robust generalization capability. $L_2$
	regularization, also called ridge regression, is widely used in full-precision
	networks. It uses squared magnitude of weights as penalty term to help avoid
	over-fitting issue. 
	
	In our binary network training approach, weight scaling factors, $\alpha$, stand
	for the actual binary filters for feature extraction. Thus, the $L_2$ regularization
	term is modified to restrict the scaling factors accordingly. The total loss
	function is then represented as
	
	\begin{equation}
	J(\alpha^l,\gamma) = L(\alpha^l,\gamma)+\frac{\lambda}{2}\sum_{l}
	||\alpha^l||_2^2
	\end{equation}
	where $L(\alpha^l,\gamma)$ is the task-related loss such as cross entropy loss.
	$\alpha^l$ is the weight scaling factors in which the superscript $l$ stands for
	different layers. $\gamma$ stands for other parameters adopted in CNN structure.
	The second term $\frac{\lambda}{2}\sum ||\alpha^l||_2^2$ is the new $L_2$
	regularization with weight decay parameter $\lambda$. The regularization term
	tends to decrease the magnitude of scaling factor $\alpha^l$. 
	
	\subsection{Compatibility with Hardware Implementation}
	\label{subsec:hardware}
	
	In this section we show that our binary network can be easily
	implemented on hardware platforms via bitwise operations. To simplify
	the discussion, we take 3x3 convolution as an example and let the input
	channel to be 1.  $\tilde{\mathcal{W}}_i\in\mathbb{R}^{3\times3}$ is
	the convolution kernel where $i$ indicates the output channel, the
	input of convolution is actually the binary activation
	$\tilde{\mathcal{A}}$ of the previous layer.  The 3$\times$3
	convolution operation is
	
	\begin{equation}
	\begin{split}
	&z=\tilde{\mathcal{A}} * \tilde{\mathcal{W}}_i = \alpha_i
	\beta(\mathcal{B}^a*\mathcal{B}^w_i) \\
	&\mathcal{B}^a = H(\mathcal{A}-\tau) \quad\in\{0,1\}\\
	&\mathcal{B}^w_i = sgn(\mathcal{W}_i) \quad\in\{-1,1\}
	\end{split}
	\end{equation}
	where $\alpha_i$ and $\beta$ are trained scaling factors for weights
	and activations respectively. For the same output channel, $\alpha_i \beta$
	is a constant so it can be integrated into the following batch
	normalization.
	
	$\mathcal{B}^a$ and $\mathcal{B}^w_i$ are both binary
	values thus $\mathcal{B}^a*\mathcal{B}^w_i$ can be implemented with
	bitwise operations. In each bit, the result of basic multiplication
	$b^a\times b^w_i$ is one of three values $\{-1,0,1\}$. If we let binary
	"$0$" stands for the value $-1$ in $b^w_i$, the truth table of binary
	multiplication is then shown in Table~\ref{tab:truthTable}, in which
	$pos$ means the result is $+1$ and $neg$ means the result is $-1$. It
	is easy to see that
	
	\begin{equation}
	pos=\mathcal{B}^a \& \mathcal{B}^w_i
	\qquad	\qquad
	neg=\mathcal{B}^a \& (\overline{\mathcal{B}^w_i})
	\end{equation}
	With $popcnt$ operation counting number of "$1$" in a binary sequence, we can
	get the binary convolution
	\begin{equation}
	\label{eq:binConv}
	\begin{split}
	\mathcal{B}^a*\mathcal{B}^w_i&=popcnt(pos)-popcnt(neg) \\
	&=popcnt(\mathcal{B}^a \&
	\mathcal{B}^w_i)-popcnt(\mathcal{B}^a \& (\overline{\mathcal{B}^w_i}))
	\end{split}
	\end{equation}
	
\begin{figure*}[!t]
		\centering
		\includegraphics[width = 4in]{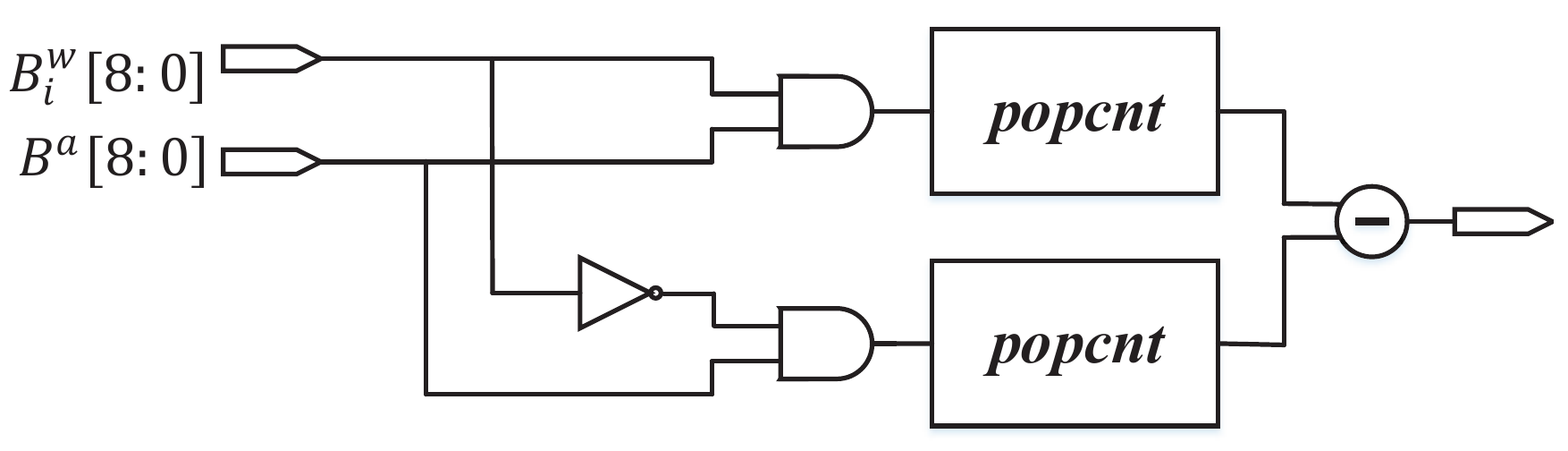}
		\caption{Hardware architecture of the binary convolution.}
		\label{fig:hardware}
	\end{figure*}

	Figure~\ref{fig:hardware} shows the hardware architecture of binary convolution
	based on Eq.~\ref{eq:binConv}. As we can see, for a 3$\times$3 convolution with
	9 multiplications and 8 additions, the binary calculation only requires 2
	$popcnt$ operations and 1 subtraction.
	
	\begin{table}[!t]
		\caption{Truth Table of Binary Multiplication $b^a\times b^w_i$.}
		\label{tab:truthTable}
		\centering
		{
			\begin{tabular}{c|c|c|c}
				\hline
				\hline
				$b^a$ & $b^w_i$ & $pos$ & $neg$ \\
				\hline
				0 & 0 & 0 & 0 \\
				\hline
				0 & 1 & 0 & 0 \\
				\hline
				1 & 0 & 0 & 1 \\
				\hline
				1 & 1 & 1 & 0 \\
				\hline
				\hline
			\end{tabular}
		}
	\end{table} 
	
	\section{Experimental Results}
	\label{sec:results}
	
	In this section, the performance of our proposed method is evaluated on image classification task. The performance of other tasks such as object detection and pose estimation will be included in the future work.
	Two image classification datasets: CIFAR-10~\cite{Krizhevsky_2009} and
	ImageNet ILSVRC12~\cite{Russakovsky_2015} are used for evaluation. The CIFAR-10 dataset
	consists of 50,000 training images with size 32$\times$32 belonging to
	10 classes and 10,000 testing images. The large ImageNet ILSVRC12
	consists of 1000 classes with about 1.2 million images for training and
	50,000 images for validation. We build the same VGG-Small network
	as~\cite{Courbariaux_2015, Hubara_2016, Rastegari_2016} for evaluation
	on CIFAR-10 dataset. For ImageNet, we implement
	AlexNet~\cite{Krizhevsky_2012} and Resnet-18~\cite{He_2016} networks.
	All networks are trained from random initialization without
	pre-training. Following previous works~\cite{Hubara_2016,
		Rastegari_2016, Zhou_2016}, we binarize all the convolution and
	fully-connected layers except the first and the last layer.  For
	regularization, we set the weight decay parameter $\lambda$ to be
	$10^{-6}$.  \cite{Hubara_2016} and~\cite{Rastegari_2016} pointed out
	that ADAM converges faster and usually performs better for binary
	inputs. Thus we use ADAM optimization for parameter updating with
	an initial learning rate $10^{-3}$ and $2\times10^{-4}$ for CIFAR-10 and
	ImageNet, respectively.
	
	\subsection{Performance on CIFAR-10}
	
	Table~\ref{tab:cifar10} presents the results of the VGG-Small network
	on CIFAR-10 dataset. The second column \emph{Bit-width} denotes
	quantization bits for weights and activations. The third column shows
	the accuracy results. Two binary network approaches,
	BNN~\cite{Hubara_2016} and XNOR-Net~\cite{Rastegari_2016}, and one
	low-precision quantization approach, HWGQ~\cite{Cai_2017}, are selected
	for comparison. The result of full-precision network model is also
	presented as a reference.  The proposed training approach achieves
	92.3\% accuracy on CIFAR-10, exceeding BNN and XNOR-Net by 2.4\% and
	2.5\%, respectively. Moreover, our result on binary network is
	very close to HWGQ~\cite{Cai_2017} with 2-bit activation quantization. 
	
	\begin{table}[!t]
		\caption{Results Comparison of VGG-Small on CIFAR-10.}
		\label{tab:cifar10}
		\centering
		{
			\begin{tabular}{c|c|c}
				\hline
				\hline
				Method & Bit-width (W/A) &Accuracy \\
				\hline
				Ours & 1/1 & \textbf{92.3\%} \\
				\hline
				BNN~\cite{Hubara_2016} & 1/1 & 89.9\% \\
				\hline
				XNOR-Net~\cite{Rastegari_2016} & 1/1 & 89.8\% \\
				\hline
				HWGQ~\cite{Cai_2017} & 1/2 & 92.5\% \\
				\hline
				Full-Precision & 32/32 & 93.6\% \\
				\hline
				\hline
			\end{tabular}
		}
	\end{table}

	\begin{figure}[!t]
		\centering
		\subfigure[Results of our approach versus
		BNN~\cite{Hubara_2016}.]{\includegraphics[width=2.5in]{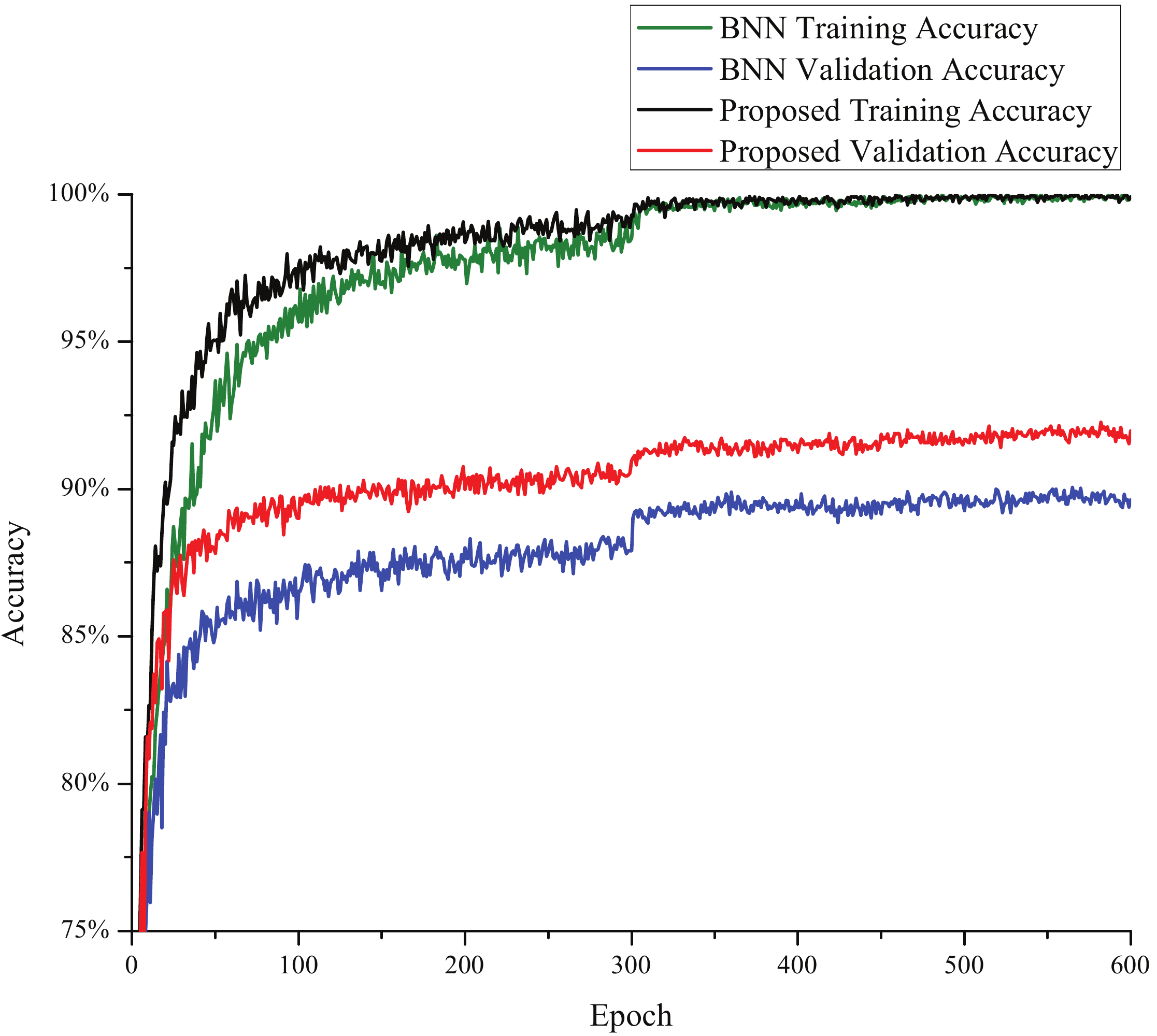}\label{fig:accuracy1}}
		\subfigure[Results with and without
		regularization.]{\includegraphics[width=2.5in]{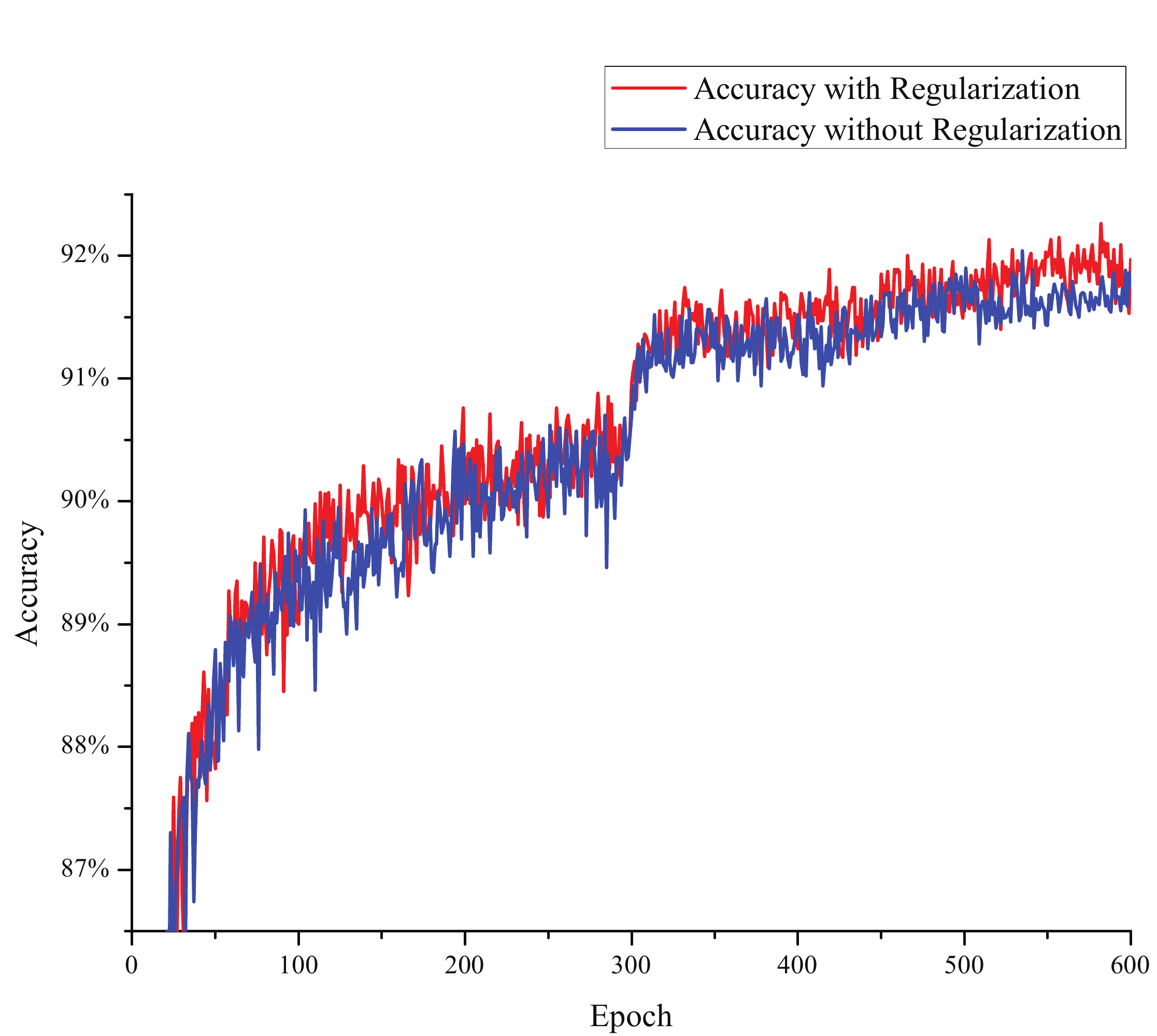}\label{fig:accuracy2}}
		\caption{Training curves comparison of VGG-Small on CIFAR-10.}
		\label{fig:accuracy}
    \end{figure}
	
	To further evaluate the proposed method, we show the training curves in
	Figure~\ref{fig:accuracy}. 
	First in Figure~\ref{fig:accuracy1}, training curves of BNN~\cite{Hubara_2016}
	and our approach are compared. Although they have similar training accuracy,
	our approach is faster to train and improves the accuracy a lot on validation
	set. Moreover, it can be observed that our validation accuracy curve is more
	stable than BNN due to the regularization. The effectiveness of regularization
	is then validated in Figure~\ref{fig:accuracy2}. The red curve is the result
	with our new $L_2$ regularization. It has less fluctuation relatively and
	the regularization term does help to improve the accuracy from 92.0\% to 92.3\%
	indicating better generalization.
		
	\subsection{Evaluations and Comparisons on ImageNet}
	
	On large ImageNet ILSVRC12 dataset, we test the accuracy performance of the
	proposed approach for AlexNet and Resnet-18 networks. The experimental results
	are presented in Table~\ref{tab:ImageNet}. We select four existing methods for
	comparison including XNOR-Net~\cite{Rastegari_2016}, BinaryNet~\cite{Tang_2017},
	ABC-Net~\cite{Lin_2017} and DoReFa-Net~\cite{Zhou_2016}. It should be noted that
	the Resnet-18 accuracy results of BinaryNet~\cite{Tang_2017} are quoted
	from~\cite{Lin_2017}. Some results of ABC-Net~\cite{Lin_2017} and
	DoReFa-Net~\cite{Zhou_2016} are not provided so we leave them blank. Similarly,
	the results of full-precision network are provided in Table~\ref{tab:ImageNet} as 
	a reference. 
	
	For AlexNet, our approach achieves 46.1\% top-1 accuracy and 70.9\% top-5
	accuracy. It is the best result among five binary network solutions and
	surpasses other works by up to 4.9\% and 5.3\%, respectively. For Resnet-18,
	our method obtains 54.2\% top-1 accuracy and 77.9\% top-5 accuracy, improving the
	performance by up to 12.0\% and 10.8\% compared with other works. It is also shown 
	that the proposed approach succeeds to reduce the accuracy gap between 
	full-precision and binary networks to about 10\%. These indicates our approach can 
	improve the inference performance of binary convolutional neural networks effectively.
	
	\begin{table}[!t]
		\caption{Results Comparisons of AlexNet and Resnet-18 on ImageNet.}
		\label{tab:ImageNet}
		\centering
		{
			\begin{tabular}{c|c|c|c|c|c}
				\hline
				\hline    
				\multirow{2}{*}{Method} & \multirow{2}{*}{Bit-width (W/A)} &
				\multicolumn{2}{c|}{AlexNet Accuracy} & \multicolumn{2}{c}{Resnet-18 Accuracy}
				\\
				\cline{3-6}
				& & Top-1 & Top-5 & Top-1 & Top-5 \\
				\hline
				Ours & 1/1 & \textbf{46.1\%} & \textbf{70.9\%} & \textbf{54.2\%} &
				\textbf{77.9\%} \\
				\hline
				XNOR-Net~\cite{Rastegari_2016} & 1/1 & 44.2\% & 69.2\% & 51.2\% & 73.2\% \\
				\hline
				BinaryNet~\cite{Tang_2017} & 1/1 & 41.2\% & 65.6\% & 42.2\% & 67.1\% \\
				\hline
				ABC-Net~\cite{Lin_2017} & 1/1 & - & - & 42.7\% & 67.6\% \\
				\hline
				DoReFa-Net~\cite{Zhou_2016} & 1/1 & 43.6\% & - & - & - \\
				\hline
				Full-Precision & 32/32 & 56.6\% & 80.2\% & 69.6\% & 89.2\% \\
				\hline
				\hline
			\end{tabular}
		}
	\end{table}

	\subsection{Analysis of Network Model Size}
	
	\begin{table}[!t]
	\caption{Network Model Size Comparison.}
	\label{tab:size}
	\centering
	{
		\begin{tabular}{c|c|c|c}
			\hline
			\hline
			\multirow{2}{*}{Network} & Full Model & Binary Model & Compression \\
			& Size & Size & Ratio \\
			\hline
			VGG-Small & 53.52MB  & 1.75MB  & 30.6$\times$ \\
			\hline
			AlexNet & 237.99MB  & 22.77MB & 10.5$\times$  \\
			\hline
			Resnet-18 & 49.83MB & 3.51MB & 14.2$\times$  \\
			\hline
			\hline
		\end{tabular}
	}
    \end{table}

	Ideally, binary neural network should achieve 32$\times$ compression ratio compared with full-precision model. But in our approach the first and the last layer are excluded from binarization operation following~\cite{Hubara_2016, Rastegari_2016, Zhou_2016}. Besides, batch normalization parameters and scaling factors should be in full precision for better network representation capability. These will affect the actual network compression ratio. Table~\ref{tab:size} shows the actual parameter size comparison of three network structures. For a typical network, binarization can reduce model size by over 10 times. Besides, it can also be observed that with a better network structure, the binary network can achieve better performance in terms of model compression ratio.

	\section{Conclusion}
	\label{sec:conclusion}
	
	This paper proposes an approach to train binary CNNs with higher
	inference accuracy including three ideas. First, trainable scaling
	factors for both weights and activations are employed to provide
	different value ranges of binary number. Then, higher-order estimator
	and long-tailed estimator for derivative of binarization function are
	proposed to balance the tight approximation and efficient
	backpropagation. At last, the $L_2$ regularization is performed
	directly on weight scaling factors for better generalization
	capability. The approach is effective to improve network accuracy and
	it is suitable for hardware implementation.
	
\bibliography{references}	
\end{document}